\documentclass{article}

\usepackage[main, final]{neurips_2026}

\usepackage{graphicx}
\usepackage{subcaption}

\usepackage[utf8]{inputenc} 
\usepackage[T1]{fontenc}    
\usepackage{hyperref}       
\usepackage{url}            
\usepackage{booktabs}       
\usepackage{amsfonts}       
\usepackage{nicefrac}       
\usepackage{microtype}      
\usepackage{xcolor}         

\usepackage{amssymb}
\usepackage{mathtools}
\usepackage{amsthm}

\usepackage{multirow}
\usepackage[normalem]{ulem}
\useunder{\uline}{\ul}{}
\usepackage[dvipsnames]{xcolor}
\usepackage{colortbl}

\usepackage{array}
\usepackage{ragged2e}   
\usepackage{placeins}
\usepackage{wrapfig}

\usepackage[capitalize,noabbrev]{cleveref}

\definecolor{defaultcolor}{gray}{.9}
\newcommand{\default}[1]{\cellcolor{defaultcolor}{#1}}

\usepackage{xspace}

\makeatletter
\DeclareRobustCommand\onedot{\futurelet\@let@token\@onedot}
\def\@onedot{\ifx\@let@token.\else.\null\fi\xspace}
\makeatother

\usepackage{pifont}
\newcommand{\eg}{\emph{e.g}\onedot}

\newcommand{\ie}{\emph{i.e}\onedot}

\newcommand{\cmark}{\ding{51}}%
\newcommand{\xmark}{\ding{55}}%
\newcommand{\ours}{UNO}

\title{Steering Visual Generation in Unified Multimodal Models with Understanding Supervision}

\author{%
    Zeyu Liu$^{1,2,3,4}$\thanks{Work done while interning at KlingAI Research.}\ \ \ \
    Zanlin Ni$^{1,2,3}$ \ \ \
    Yang Yue$^{1,2,3}$ \ \ \
    Cheng Da$^{4}$ \ \ \
    Huan Yang$^{4}$ \\
    \textbf{Di Zhang$^{4}$} \ \ \
    \textbf{Kun Gai$^{4}$} \ \ \
    \textbf{Gao Huang$^{1,2,3}$\thanks{Corresponding author.}} \\
    $^{1}$Department of Automation, Tsinghua University \\ $^{2}$Beijing Key Laboratory of Embodied Intelligence Systems  \\
    $^{3}$Institute for Embodied Intelligence and Robotics, Tsinghua University \\
    $^{4}$KlingAI Research \\
    Project Page: \url{https://lzy-tony.github.io/uno}
}

\begin{document}

\maketitle

\begin{abstract}

Unified multimodal models are envisioned to bridge the gap between understanding and generation. 
Yet, to achieve competitive performance, state-of-the-art models adopt largely decoupled understanding and generation components.
This design, while effective for individual tasks, weakens the connection required for mutual enhancement, leaving the potential synergy empirically uncertain.
We propose to explicitly restore this synergy by introducing \emph{\textbf{Un}derstanding-\textbf{O}riented Post-Training} (\textbf{UNO}), a lightweight framework that treats understanding not only as a distinct task, but also a direct supervisory signal to steer generative representations.
By incorporating objectives that encode semantic abstraction (captioning) and structural details (visual regression), we enable effective gradient flow from understanding to generation.
Extensive experiments on image generation and editing demonstrate that understanding can serve as an effective catalyst for generation.

\end{abstract}    
\section{Introduction}
\label{sec:intro}

Unified Multimodal Models (UMMs), which integrate language comprehension, visual understanding, and visual generation within a single framework, have recently achieved significant success~\cite{team2024chameleon, dongdreamllm, wang2024emu3, tong2024metamorph, pan2025transfer, deng2025emerging, chen2025blip3}. By jointly modeling understanding and generation, these models facilitate versatile any-to-any interaction, enabling advanced and promising new capabilities that range from complex multimodal reasoning~\cite{liimagine, chern2025thinking} and free-form image manipulation~\cite{deng2025emerging} to interleaved world modeling~\cite{gou2025vq, wu2026visual}.

A long-term objective for Unified Multimodal Models is to achieve capability synergy~\cite{dongdreamllm, tong2024metamorph, wu2026liquid}, wherein multimodal understanding and generation do not merely coexist under a single framework, but mutually enhance one another. However, to maintain strong task-specific performance, state-of-the-art architectures increasingly adopt a decoupled representation paradigm~\cite{liang2024mixture, wu2025janus, chen2025janus, ma2025janusflow, deng2025emerging}, which aims to alleviate optimization conflicts between the high-level semantic abstraction required for understanding and the low-level objectives inherent to generative modeling. Concretely, these approaches separate understanding and generation into distinct representation spaces, ranging from distinct vision encoders~\cite{wu2025janus} and feed-forward networks (FFNs)~\cite{li2025onecat} to disjoint transformer parameters~\cite{liang2024mixture, deng2025emerging}. While such decoupling effectively mitigates interference and preserves specialization, it inherently limits the extent to which the rich semantics learned by the understanding expert can be directly utilized by the generative components, leaving it less clear to what extent genuine capability synergy can be achieved within these frameworks.

\begin{figure*}[t]
    \centering
    \includegraphics[width=\linewidth]{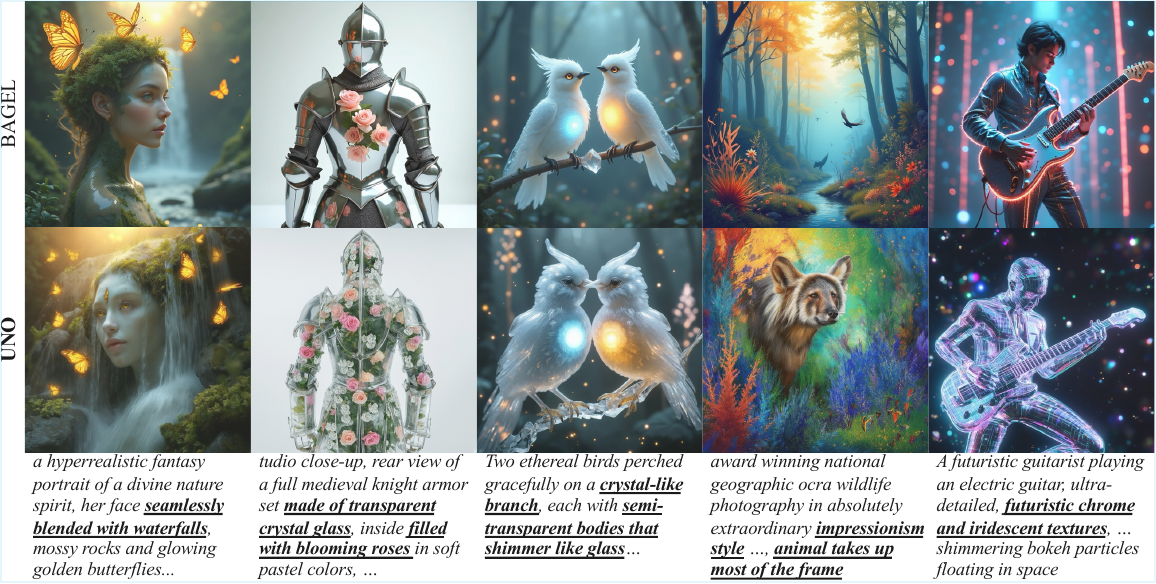}
    \caption{\textbf{Qualitative comparisons on image generation} between BAGEL and BAGEL-\ours. }
    \label{fig:cmp_gen}
    \vspace{-6mm}
\end{figure*}

In this work, we isolate this specific direction of synergy and investigate whether direct understanding-oriented supervision can be systematically leveraged to enhance generative learning in unified models. To this end, we propose \emph{\textbf{Un}derstanding-\textbf{O}riented Post-Training} (\textbf{UNO}), a light-weight framework that explicitly supervises generative representations with understanding signals. Rather than treating understanding as a parallel task, we re-route the information flow by conditioning the frozen understanding expert on intermediate noised generative representations, strengthening direct gradient flow from understanding to generation. Specifically, we incorporate two complementary understanding-oriented proxy objectives for optimizing generative representations: (i) language supervision via captioning and (ii) visual understanding supervision via regressing with metaquery tokens. Language supervision enhances discriminative concepts through high-level abstraction, yet is inherently sparse and may overlook fine-grained details. In contrast, visual understanding supervision captures denser details and spatial structures, providing the structural information that abstract linguistic signals lack. By integrating these complementary objectives, \ours \ enriches the model's generative representations with multimodal semantics without increased architectural complexity.

Building on this conceptual framework, we conduct a systematic evaluation across diverse generation tasks. Extensive experiments across image generation and editing benchmarks with two different backbones indicate that \ours \  yields consistent and substantial improvements over strong baselines without degrading understanding performance. Specifically, \ours \ improves the competitive BAGEL~\cite{deng2025emerging} baseline on both image generation (GenEval2  71.7 $\rightarrow$ 75.1, DPG-Bench 84.03 $\rightarrow$ 86.12, UniGenBench++ 61.53 $\rightarrow$ 65.03) and image editing tasks (GEdit-Bench-EN 6.52 $\rightarrow$ 7.17, GEdit-Bench-CN 6.50 $\rightarrow$ 7.20) by significant margins. Beyond quantitative gains, qualitative visualizations further reveal improved semantic structure for generative representations at heavily noised timesteps. These results demonstrate that in unified models, strong multimodal understanding can be harnessed to directly benefit generation, paving the way for more integrated unified multimodal systems.

\section{Related Work}
\label{sec:related_work}

\begin{figure*}
    \centering
    \includegraphics[width=\linewidth]{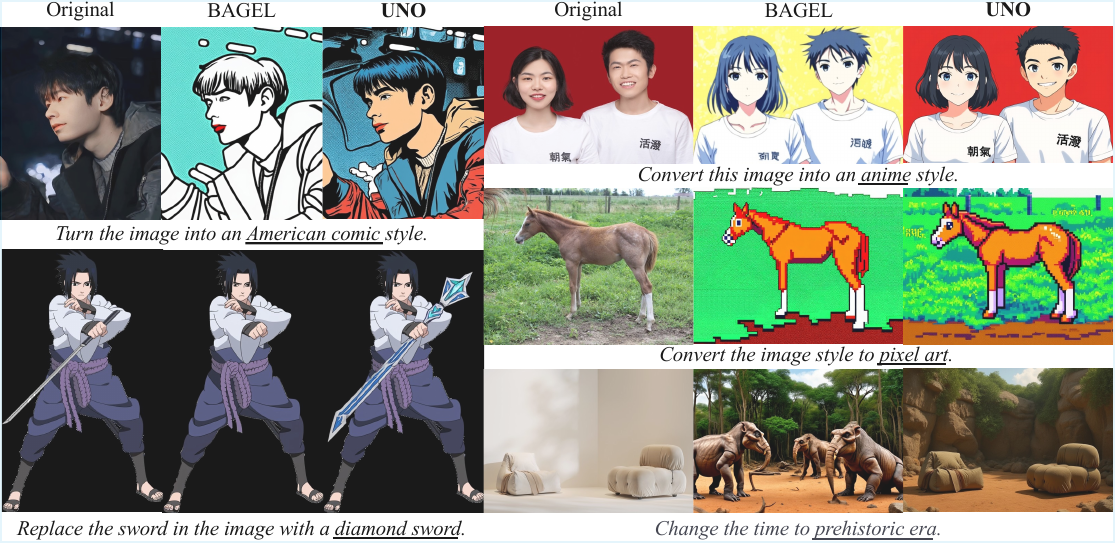}
    \caption{\textbf{Qualitative comparisons on image editing} between BAGEL and BAGEL-\ours. }
    \label{fig:cmp_edit}
    \vspace{-6mm}
\end{figure*}

\subsection{Unified Multimodal Models}

Inspired by the success of large language models (LLMs)~\cite{achiam2023gpt, touvron2023llama, guo2025deepseek, yang2025qwen3} and advances in separate multimodal understanding~\cite{liu2023visual, wang2024qwen2, bai2025qwen2} and generation~\cite{rombach2022high, podellsdxl, peebles2023scalable, esser2024scaling, flux} systems, recent works have moved toward unified multimodal models that perform both multimodal understanding and generation within a unified framework. Early approaches often relied on quantized autoregressive modeling of visual content~\cite{team2024chameleon, wang2024emu3, wu2025janus, chen2025janus, wu2024vila}, \ie, transforming images into a sequence of tokens with discrete vector quantizers and processing those tokens autoregressively in a way akin to language modeling. While these methods demonstrated the feasibility of unified modeling, their generative quality is constrained by the discretization bottleneck imposed by VQ-based tokenizers. To overcome these limitations, recent approaches combine multimodal large language models (MLLMs) for understanding with diffusion models for generation, yielding substantially improved expressivity and performance. Within this hybrid paradigm, one thread of research arranges an MLLM backbone sequentially with a diffusion decoder. Implementations include either predicting through special query tokens~\cite{sun2024generative, dongdreamllm, ge2024seed, pan2025transfer, wu2025openuni}, or through predicting intermediate latent representations ~\cite{tong2024metamorph, chen2025blip3} that are consumed by the diffusion-based generator. A complementary line of work emphasizes parallel architectures that process understanding and generation within a unified backbone. These designs include integrated transformer~\cite{ma2025janusflow, xie2025show} as well as Mixture-of-Experts \cite{li2025onecat} or Mixture-of-Transformers \cite{liang2024mixture, liao2025mogao, deng2025emerging}  formulations that allocate capacity across modalities and tasks.

\subsection{Representations in Unified Multimodal Models}

A central challenge in unified multimodal modeling lies in representation design. Unified models must simultaneously support multiple, potentially conflicting tasks, each imposing distinct requirements on the underlying representations. While early unified models enforced a single shared representation for all visual signals~\cite{team2024chameleon, wang2024emu3}, subsequent studies have shown that such designs lead to potential task conflicts~\cite{xieshow} that degrade task-specific performance. As a result, contemporary models increasingly adopt \emph{decoupled visual representations} to better accommodate divergent objectives. One common strategy is to employ separate vision encoders for understanding and generation~\cite{wu2025janus, chen2025janus, xie2025show}. Beyond encoder decoupling, more recent architectures further separate representations within the backbone itself. Mixture-based designs, including MoE and MoT~\cite{liang2024mixture, liao2025mogao, deng2025emerging, li2025onecat}, explicitly allocate distinct pathways to understanding and generation. This suggests that unified models operate over multiple representations and that effective coordination among these representations is critical for performance.

\subsection{Understanding Priors for Generation}

Recent studies have increasingly highlighted the importance of understanding-oriented priors in improving generative modeling. Representation alignment methods such as REPA~\cite{yurepresentation} and REPA-E~\cite{leng2025repa} regularize diffusion training by aligning intermediate features with pretrained semantic visual representations, substantially accelerating convergence and performance. Beyond alignment losses, RAE~\cite{zheng2025diffusion, tong2026scaling} and SVG~\cite{shi2025latent, shi2025svg} redesign latent spaces around semantically rich encoder representations, enabling high-quality generation without relying on traditional VAEs. 

\section{Approach}
\label{sec:approach}

\begin{figure*}[t]
    \centering
    \includegraphics[width=\linewidth]{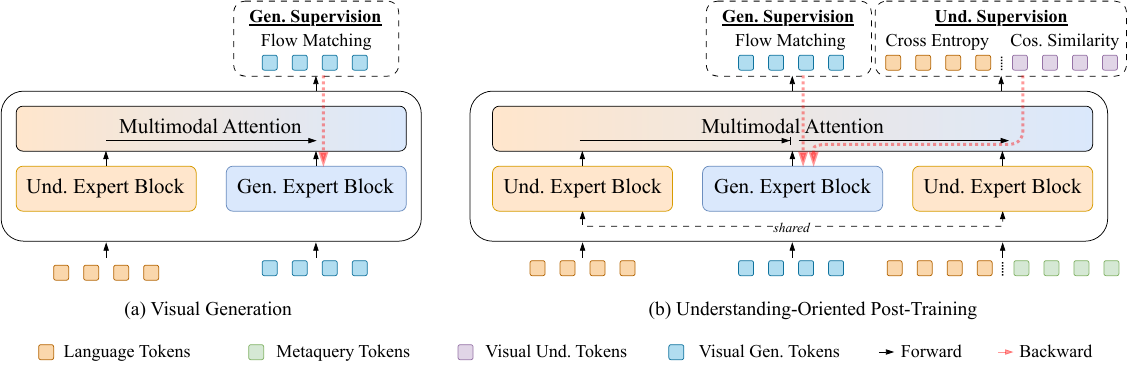}
    \caption{\textbf{Conceptual illustration of training process and backward gradient flow}. (a) \textit{Generation Training}: Current generative training in unified models encode conditions using the understanding expert and transfers information uni-directionally via conditioning to the generative expert, where the outputs are optimized using low-level flow matching objectives. Generation experts receive gradients solely from generative targets, while signals from the understanding pathway remain isolated. (b) \textit{Understanding-Oriented Post-Training}: Understanding-oriented post-training jointly supervises the sample with generation and understanding objectives. Specifically, the understanding expert conditions on the noised generative representations, and jointly conducts (i) language supervision through captioning and (ii) visual understanding supervision by using metaquery tokens to predict subsequent understanding tokens. This enables generation blocks to receive gradients from both pathways, enabling strong understanding supervision to directly shape generative representations.}
    \label{fig:pipeline}
    \vspace{-4mm}
\end{figure*}

\subsection{Preliminary: Information Flow and Representations in Unified Multimodal Models} 

Representative state-of-the-art unified multimodal models, \eg BAGEL~\cite{deng2025emerging}, are typically initialized from pretrained vision–language models (VLMs) and comprise specialized experts for understanding and generation. Visual understanding is handled exclusively by the understanding expert, which jointly processes visual understanding and language tokens in isolation from the generation pathway. Conversely, visual generation is conditioned on representations encoded by the understanding expert and supervised by low-level flow-matching objectives, as illustrated in \cref{fig:pipeline}(a). As a result, although unified models consolidate diverse capabilities within a single architecture, their internal representations adopt decoupled designs and exhibit distinct characteristics, as summarized in \cref{tab:features}. Language representations are highly abstract but lack dense information arising from visual details; conversely, generation representations are visually dense but often lack rigorous semantic organization. Visual understanding representations occupy an intermediary position, retaining visual granularity while encoding structured semantics. 

\subsection{Motivation}

\begin{wraptable}{r}{0.34\textwidth}
    \centering
    \footnotesize
    \vspace{-4mm}
    \caption{\textbf{Different representations} in UMMs (\eg BAGEL~\cite{deng2025emerging}).}
    \label{tab:features}
    \setlength{\tabcolsep}{1.5pt}
    \begin{tabular}{lll}
    \toprule
    Representation & Abstraction & Density \\
    \midrule
    Language & High & Low \\
    Visual Und. & Medium & Medium \\
    Visual Gen. & Low & High \\
    \bottomrule
    \end{tabular}
    \vspace{-4mm}
\end{wraptable}

As previously established, the decoupled architecture induces a unidirectional information flow from understanding to generation. Although the generation expert is conditioned on the understanding expert and can therefore inherit semantic cues implicitly, the generative flow-matching objective provides only weak supervision for enforcing semantic structure in the generative representations. Consequently, a pronounced performance gap emerges: while the understanding expert exhibits strong semantic capabilities, the generation expert frequently struggles with complex instructions and fine-grained semantic adherence. This disparity indicates that, in current unified frameworks, understanding capabilities are substantially stronger than generation, yet remain largely under-exploited as a source of supervision. Importantly, unified models already encode rich semantic representations through language and visual understanding, therefore we argue that relying solely on low-level objectives is suboptimal for training the generative pathway. Motivated by this observation, we hypothesize that explicitly supervising generative representations through understanding objectives operated by the model's own understanding expert can inject strong semantic constraints, yielding a more semantically grounded representation space and ultimately improved generative performance.

\subsection{Language and Visual Understanding Supervision}

\noindent\textbf{Language Supervision} To address this, we first exploit the strong language-based understanding capabilities and re-route the information flow to enable \textit{language supervision} directly over generated visual representations, as conceptually illustrated in ~\cref{fig:pipeline}(b). Rather than conditioning the understanding expert on visual understanding tokens, we condition it on the noised generation representations from the generation expert. The understanding expert then processes these features to output language tokens, supervised by an image captioning objective. This forces the understanding expert to decode semantics directly from the intermediate generation representations, effectively distilling abstract pretrained semantics into the generative pathway through backward gradient flows from understanding to generation. To preserve pretrained capabilities, we freeze the understanding expert, compelling the generation representations to adapt to the understanding-oriented objective.

\begin{wrapfigure}{r}{0.28\textwidth}
    \centering
    \vspace{-4mm}
    \includegraphics[width=\linewidth]{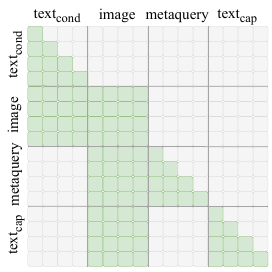}
    \caption{\textbf{Attention mask} for packed text-to-image training sample sequence.}
    \label{fig:attention_mask}
    \vspace{-4mm}
\end{wrapfigure}

A critical challenge in this setup is avoiding trivial solutions from information leakage. To mitigate this, we mask conditional prompt tokens when forwarding supervision language tokens, as shown in ~\cref{fig:attention_mask}. However, we empirically observe that supervising generated images with captions derived from the original prompt leads to abnormally low captioning loss. We hypothesize that the high information capacity of visual representations allows the model to ``store" low-density prompt signals, creating shortcuts where the understanding expert simply copies rather than performing genuine semantic extraction.

To mitigate this, we adopt semantic augmentations by re-captioning the images using alternative captioning models. Target captions are semantically consistent but lexically different from the original conditioning prompts. By supervising the model with non-token-aligned captions, we force the understanding expert to rely on extracting semantic content from generation representations rather than surface-level token copying. The resulting objective is defined as:
\begin{equation}
    \mathcal L_{\mathrm{language}} = -\sum \limits_{i = 1}^L \log p(\mathbf z_i | \mathbf z_{< i}, \mathbf V_{\mathrm{gen}}),
\end{equation}
where $\mathbf z$ denotes the supervision tokens, and $\mathbf V_{\mathrm{gen}}$ represents the noised visual representations.

\noindent\textbf{Visual Understanding Supervision} While language supervision provides high-level semantic guidance, language-based captions are inherently limited in information density: they often omit fine-grained visual details and cannot fully describe all aspects of an image. Moreover, language supervision lacks explicit 2D semantic structure encoded in visual based representations. To address these limitations, we introduce \emph{visual understanding supervision} that complement language-based signals.

Specifically, we use the pretrained understanding expert as a strong visual prior. As the understanding expert is not designed to directly produce visual features, we adopt the MetaQuery framework~\cite{pan2025transfer} and insert a set of learnable metaqueries into the understanding expert. These metaqueries are processed autoregressively, and their output hidden states are supervised to regress dense visual features extracted from the target image by the model's native visual encoder~\cite{tschannen2025siglip}. This yields the visual supervision objective:
\begin{equation}
    \mathcal L_{\mathrm{vision}} = - \sum \limits_{j = 1}^N \mathrm{sim} (\mathbf{v}_j, \mathbf h_j), 
\end{equation}
where $\mathbf{v}_j$ and $\mathbf{h}_j$ denotes the target representations from the pretrained visual encoder and the output states of the metaqueries, respectively. $\mathrm{sim}(\cdot)$ represents cosine similarity.

\noindent\textbf{Joint Supervision} As summarized in~\cref{tab:features}, visual understanding representations contains fine-grained details and explicit 2D spatial structure that enrich language supervision. Conversely, language captions offer a more abstract and direct supervision, providing complementary high-level semantic guidance to visual supervision. Together, these two forms of supervision enable a more comprehensive learning signal. To enable joint supervision, we combine the proposed objectives with the standard flow-matching loss:
\begin{equation}
    \mathcal L_{\mathrm{total}} = \mathcal L_{\mathrm{mse}} + \lambda_1 \mathcal L_{\mathrm{language}} + \lambda_2 \mathcal L_{\mathrm{vision}},
\end{equation}
To maximize training data efficiency, we employ a unified data packing strategy that concatenates all supervision signals into a single sequence. As shown in \cref{fig:attention_mask}, we modify the attention mask to manage information flow and avoid leakage between these tasks. This end-to-end configuration forces the generative pathway to optimize for both generation and understanding signals simultaneously.

\section{Experiments}
\label{sec:exp}

\begin{table*}[t]
    \caption{\textbf{Quantitative image generation performance}. $^\dagger$: evaluation with self-CoT. $^*$: locally evaluated results. Know.: World knowledge. Attr.: Attribute. Act.: Action. Relat.: Relationship. Rea.: Logical reasoning. Gram.: Grammar. Comp.: Compound. Lay.: Layout. Ovr.: Overall.}
    \label{tab:main}
    \centering
    \footnotesize
    \setlength{\tabcolsep}{1pt}
    \begin{tabular}{clccccccccccccc}
    \toprule
     & \multirow{2}{*}{Model} & \multirow{2}{*}{GenEval2} & \multirow{2}{*}{DPG-Bench} & \multicolumn{11}{l}{\ \ \ \ \ \ \ \ \ \ \ \ \ \ \ \ \ \ \ \ \ \ \ \ \ \ \ \ \ \ \ \ \ \ \ \ \ \ \ \ \ \ \ \ \ \ \ \ \ \  UniGenBench++} \\
    \cmidrule{5-15}
     & & & & Style  & Know.  & Attr.  & Act.   & Relat. & Rea. & Gram. & Comp.  & Lay. & Text  & \textbf{Ovr.} \\
    \midrule
    \multirow{7}{*}{{\ul \textit{Gen. Only}}} & SDXL & 45.3 & 74.65 & 87.40 & 72.63 & 44.34 & 34.22 & 44.92 & 9.55 & 47.33 & 26.68 & 29.85 & 1.15 & 39.75  \\
    & SD-3.5-Medium & 64.8$^*$ & 83.79$^*$ & 89.80 & 84.34 & 66.99 & 60.65 & 68.78 & 37.73 & 59.89 & 53.35 & 70.34 & 15.23 & 60.71 \\
    & SD-3.5-Large & 67.8 & 83.86$^*$ & 88.60 & 88.92 & 68.59 & 62.17 & 69.80 & 32.27 & 58.96 & 58.76 & 69.03 & 32.76 & 62.99  \\
    & FLUX.1-dev & 67.0 & 84.00 & 83.90 & 88.92 & 67.84 & 62.17 & 67.26 & 30.91 & 60.96 & 47.04 & 71.83 & 32.18 & 61.39 \\
    & Infinity & 61.2$^*$ & 83.46 & 90.80 & 87.97 & 68.06 & 60.17 & 69.16 & 31.36 & 60.16 & 51.42 & 66.60 & 12.36 & 59.81 \\
    & OmniGen2 & 67.4$^*$ & 83.57 & 91.90 & 86.39 & 72.12 & 62.83 & 68.27 & 32.50 & 59.89 & 56.31 & 71.64 & 29.02 & 63.09 \\
    & Wan2.2-t2i-plus & - & - & 91.10 & 87.34 & 70.19 & 68.00 & 73.03 & 42.05 & 66.53 & 61.37 & 74.77 & 13.83 & 64.82 \\
    \midrule
    \multirow{10}{*}{{\ul \textit{Unified}}}   & Janus & 53.7$^*$ & 79.68 & 89.90 & 73.58 & 54.81 & 50.38 & 55.08 & 26.82 & 59.09 & 46.65 & 54.85 & 1.15 & 51.23 \\
     & Janus-Pro-7B & 54.5$^*$ & 84.19 & 90.80 & 86.71 & 67.74 & 64.26 & 68.40 & 37.05 & 64.44 & 62.11 & 72.01 & 2.59 & 61.61 \\
     & Emu3-8B & 41.2$^*$ & 80.60 & 86.80 & 77.06 & 51.39 & 40.11 & 49.75 & 19.32 & 52.94 & 36.86 & 44.78 & 1.15 & 46.02 \\
     & OneCAT-3B & 60.7$^*$ & 84.53 & 93.30 & 82.28 & 63.46 & 58.56 & 68.15 & 33.41 & 60.83 & 56.96 & 64.74 & 1.15 & 58.28 \\
     & Janus-Flow & 42.7$^*$ & - & 86.20 & 62.50 & 47.97 & 43.35 & 50.00 & 21.14 & 60.29 & 45.10 & 46.46 & 0.86 & 46.39 \\
     & BLIP3-o-8B & 59.8$^*$ & 81.60 & 92.80  & 80.22  & 63.89  & 63.97  & 66.50  & 39.55   & 68.58   & 53.74  & 68.47  & 1.15   & 59.87   \\
     & UniWorld-V1 & 70.0$^*$ & - & 91.10 & 82.91 & 70.62 & 67.21 & 67.13 & 38.41 & 63.77 & 54.51 & 69.03 & 26.44 & 63.11  \\
     & Mogao-7B & - & 84.33 & -  & -  & -  & -  & -  & -   & -   & -  & -  & -   & -   \\
     & BAGEL & 71.7$^\dagger$ & 84.03 & 90.20  & 85.60  & 67.74  & 61.98  & 70.69  & 30.23   & 66.44   & 58.12  & 76.49  & 7.76   & 61.53   \\
     & \textbf{BAGEL + \ours} & \textbf{75.1}$^\dagger$ & \textbf{86.12} & 92.30 & 84.18 & 72.86 & 67.49 & 75.25 & 33.26 & 68.98 & 64.82 & 80.78 & 10.34 & \textbf{65.03} \\
     & $\Delta$ & \textbf{\textcolor{ForestGreen}{+3.4}} & \textbf{\textcolor{ForestGreen}{+2.09}} & \textbf{\textcolor{ForestGreen}{+2.10}} & \textbf{\textcolor{gray}{-1.42}} & \textbf{\textcolor{ForestGreen}{+5.12}} & \textbf{\textcolor{ForestGreen}{+5.51}} & \textbf{\textcolor{ForestGreen}{+4.56}} & \textbf{\textcolor{ForestGreen}{+3.03}} & \textbf{\textcolor{ForestGreen}{+2.54}} & \textbf{\textcolor{ForestGreen}{+6.70}} & \textbf{\textcolor{ForestGreen}{+4.29}} & \textbf{\textcolor{ForestGreen}{+2.58}} & \textbf{\textcolor{ForestGreen}{+3.50}} \\
     \bottomrule
    \end{tabular}
    \vspace{-4mm}
\end{table*}

\subsection{Experiment Setup}

\noindent{\textbf{Training}} For all experiments, we train BAGEL-7B~\cite{deng2025emerging} for 5K iterations while keeping the understanding expert frozen. For the image generation task, we utilize a curated set of high-quality text-image pairs. Notably, we exclude distillation-based data such as BLIP3o-60k~\cite{chen2025blip3} to prevent evaluation template leakage for certain evaluation benchmarks~\cite{xie2025reconstruction}. We apply semantic augmentation and re-caption images using different caption models to form text-image-text triplets. For image editing, training is conducted on CrispEdit-2M~\cite{chow2025editmgt}, a diverse set of high quality editing pairs. We additionally caption the target images to provide supervision text, resulting in text-image-image-text quadruplets.

\noindent{\textbf{Evaluation Protocol}} We evaluate text-to-image generation performance using GenEval2~\cite{kamath2025geneval}, DPG-Bench~\cite{hu2024ella} and UniGenBench++~\cite{wang2025unigenbench++}. We primarily focus on DPG-Bench for its diverse prompts to evaluate semantic related instruction following. Additionally, as DPG-Bench exhibit rapid performance saturation~\cite{tang2025exploring}, we also evaluate on UniGenBench++, a more recent and fine-grained evaluation set. We also report results on GenEval2~\cite{kamath2025geneval}, an improved version of GenEval~\cite{ghosh2023geneval} that mitigates evaluation errors and drifts from human judgment. Evaluations on DPG-Bench and UniGenBench++ are conducted without activating thinking mode or prompt rewriting, while GenEval2 is evaluated with CoT following default setting in ~\cite{kamath2025geneval}. We evaluate with 4 random seeds to balance robustness and computation costs. For image editing, we evaluate on GEdit-Bench-EN/CN~\cite{liu2025step1x}, a comprehensive multilingual benchmark derived from real-world user instructions.

\begin{figure*}[t]
    \centering
    \includegraphics[width=\linewidth]{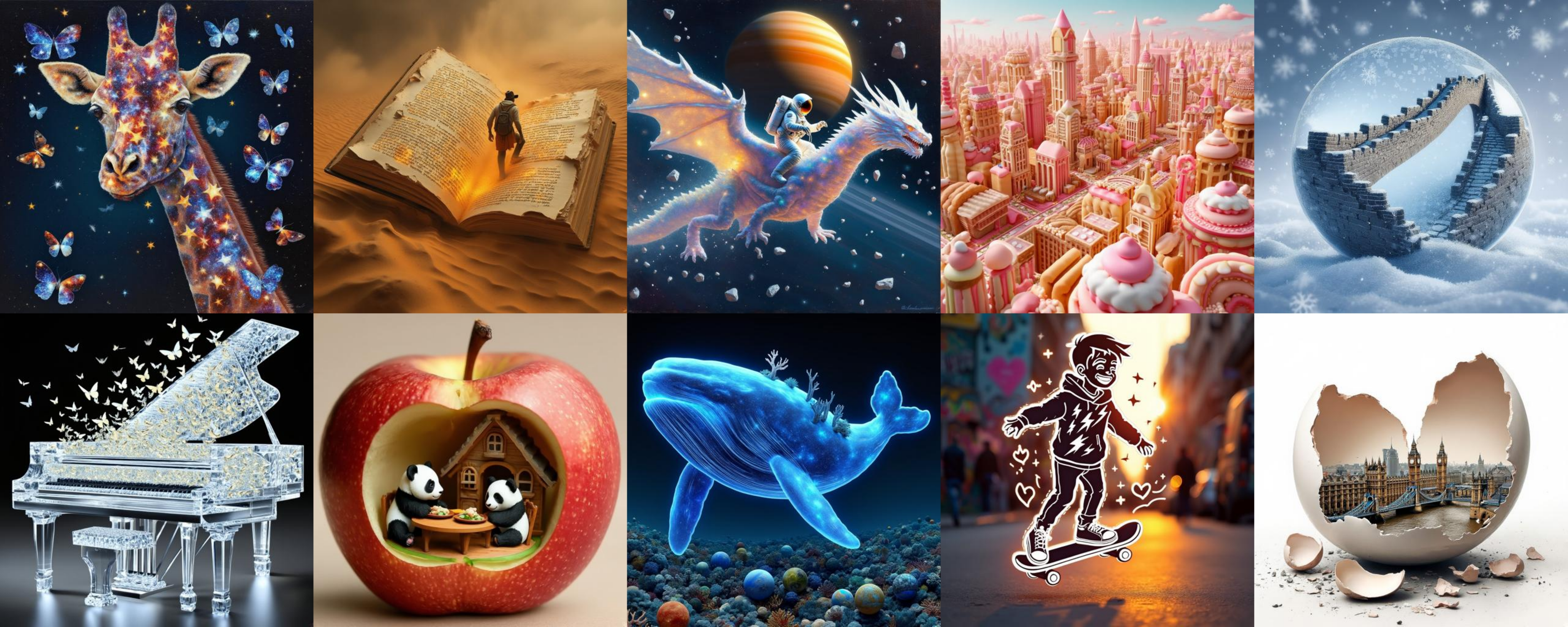}
    \caption{Qualitative visualizations of image generation results.}
    \label{fig:visualize_gen}
    \vspace{-4mm}
\end{figure*}

\noindent{\textbf{Baselines}} We compare against both generation-only and unified models. For image generation, generation-only baselines include SDXL~\cite{podellsdxl}, Stable Diffusion 3.5 Medium/Large~\cite{esser2024scaling}, FLUX.1-dev~\cite{flux}, Infinity~\cite{han2025infinity}, OmniGen2~\cite{wu2025omnigen2} and Wan2.2-t2i-plus, unified models include Janus~\cite{wu2025janus}, Janus-Pro~\cite{chen2025janus}, Emu3~\cite{wang2024emu3}, OneCAT~\cite{li2025onecat}, Janus-Flow~\cite{ma2025janusflow}, BLIP3-o~\cite{chen2025blip3}, UniWorld-V1~\cite{lin2025uniworld}, Mogao~\cite{liao2025mogao} and BAGEL~\cite{deng2025emerging}. For editing, generation-only models include Instruct-Pix2Pix~\cite{brooks2023instructpix2pix}, MagicBrush~\cite{zhang2023magicbrush}, AnyEdit~\cite{jianganyedit}, OmniGen~\cite{xiao2025omnigen}, OmniGen2~\cite{wu2025omnigen2}, Step1X-Edit~\cite{liu2025step1x} and FLUX-Kontext~\cite{labs2025flux}, and unified models include BAGEL~\cite{deng2025emerging}, BAGEL-NHR~\cite{kuprashevich2025nohumansrequired} and UniWorld-V1~\cite{lin2025uniworld}.

\subsection{Main Results}

\noindent\textbf{Image Generation} We report generation performance in~\cref{tab:main}. \ours \  yields consistent improvements over the original BAGEL as well as generation-only and unified-model baselines across GenEval2, DPG-Bench, and UniGenBench++. On UniGenBench++, \ours \  shows pronounced gains in dimensions including compound, attribute, action, and relationship, which are closely related to semantic comprehension. We observe only a slight degradation in world-knowledge scores, indicating that it is more dependent on the diversity and coverage of the training data.

\begin{wraptable}{r}{0.65\textwidth}
    \centering
    \footnotesize
    \vspace{-4mm}
    \setlength{\tabcolsep}{2.5pt}
    \caption{\textbf{Quantitative image editing performance}. G\_SC, G\_PQ and G\_O denotes GPT-4.1 evaluated semantic consistency, perceptual quality and overall performance. We mark the best results for each dimension in \textbf{bold} and {\ul underline} the second best.}
    \label{tab:edit}
    \begin{tabular}{lcccccc}
    \toprule
    \multirow{2}{*}{Model} & \multicolumn{3}{l}{\ \, GEdit-Bench-EN} & \multicolumn{3}{l}{\ \, GEdit-Bench-CN} \\
    & G\_SC      & G\_PQ     & \textbf{G\_O} & G\_SC      & G\_PQ     & \textbf{G\_O}     \\
    \midrule
    Instruct-Pix2Pix & 3.58 & 5.49 & 3.68 & - & - & - \\
    MagicBrush & 4.68 & 5.66 & 4.52 & - & - & -\\
    AnyEdit & 3.18 & 5.82 & 3.21 & - & - & -\\
    OmniGen & 5.96 & 5.89 & 5.06 & - & - & -\\
    OmniGen2 & 7.16 & 6.77 & 6.41 & - & - & -\\
    Step1X-Edit & 7.09 & 6.76 & 6.70 & 7.20 & {\ul 6.87} & {\ul 6.86} \\
    FLUX-Kontext & 6.95 & {\ul 7.30} & 6.27 & - & - & -\\
    UniWorld-V1 & 4.93 & 7.43 & 4.85 & - & - & - \\
    BAGEL-NHR & \textbf{8.04} & 6.87 & {\ul 7.08} & - & - & - \\
    BAGEL & 7.36 & 6.83 & 6.52 & {\ul 7.34} & 6.85 & 6.50 \\
    \textbf{BAGEL + \ours} & {\ul 7.76} & \textbf{7.54} & \textbf{7.17} & \textbf{7.80} & \textbf{7.54} & \textbf{7.20} \\
    $\Delta$ & \textbf{\textcolor{ForestGreen}{+0.40}} & \textbf{\textcolor{ForestGreen}{+0.71}} & \textbf{\textcolor{ForestGreen}{+0.65}} & \textbf{\textcolor{ForestGreen}{+0.46}} & \textbf{\textcolor{ForestGreen}{+0.69}} & \textbf{\textcolor{ForestGreen}{+0.70}} \\
    \bottomrule
    \end{tabular}
    \vspace{-6mm}
\end{wraptable}

\noindent\textbf{Image Editing} We report quantitative image editing results on GEdit-Bench-EN/CN in~\cref{tab:edit}. \ours \  consistently improves editing performance over BAGEL and other strong baselines across semantic consistency, perceptual quality, and overall metrics. On GEdit-Bench-EN, \ours \  achieves the best overall score, with notable gains in perceptual quality while preserving edit intent. Improvements also transfer to Chinese evaluations on GEdit-Bench-CN despite training solely on English data, demonstrating robust generalization.

\noindent\textbf{Qualitative Results.} We present qualitative comparisons of image generation and editing between the original BAGEL and \ours \  in~\cref{fig:cmp_gen} and~\cref{fig:cmp_edit}, respectively. A more comprehensive comparison is presented in \cref{sec:more_visualization} of the appendix. For generation, \ours \  demonstrates stronger instruction following. For editing, it more effectively interprets abstract instructions and better preserves fine-grained background details, benefiting from the additional understanding supervision during training. Further qualitative examples are provided in~\cref{fig:visualize_gen} and~\cref{fig:visualize_edit}. Full prompt list is displayed in \cref{tab:full_prompt} of the appendix.

\begin{figure*}[t]
    \centering
    \includegraphics[width=\linewidth]{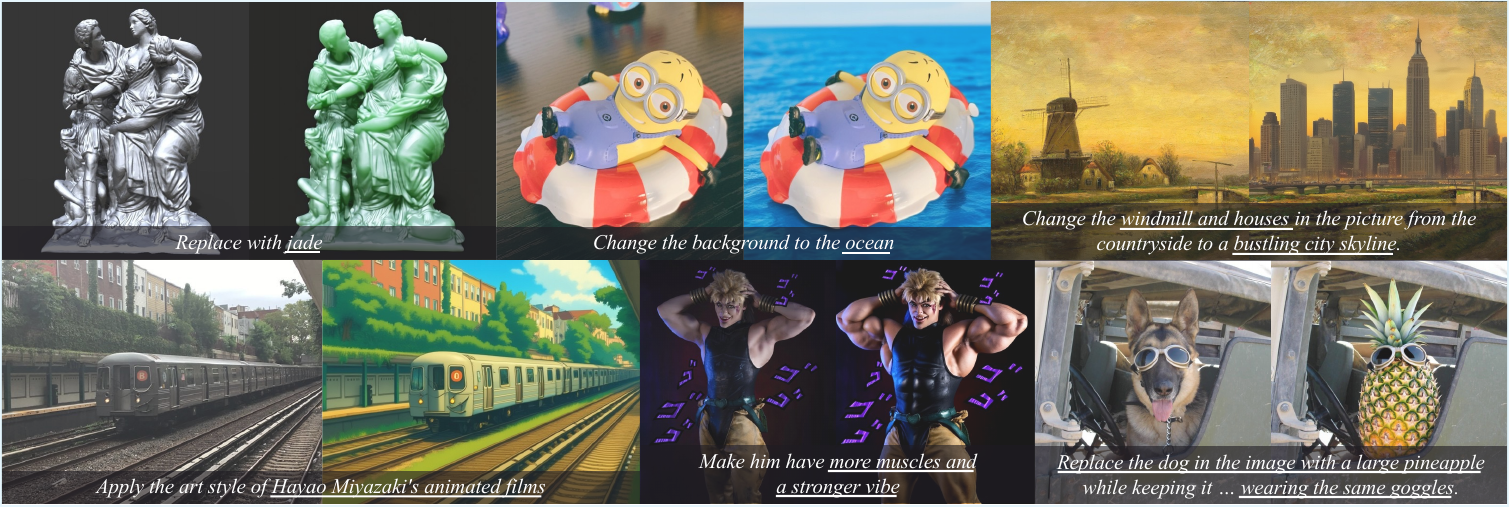}
    \caption{Qualitative visualizations of image editing results.}
    \label{fig:visualize_edit}
    \vspace{-2mm}
\end{figure*}

\subsection{Analysis}

\noindent\textbf{\ours \  as an effective post-training approach} To assess the effectiveness of \ours \  as a post-training strategy, we decompose \ours \  and compare it against other tuning-based post-training approaches under identical data settings. For image generation, we compare with supervised fine-tuning (SFT) and reconstruction alignment (RecA~\cite{xie2025reconstruction}), and present results in ~\cref{tab:post_train}. Our observations are threefold: 1) applying either language or visual understanding supervision alone consistently improves performance over SFT, with language supervision exhibiting most pronounced gains. 2) Joint supervision yields further gains beyond single-modality supervision, indicating a complementary relationship language and visual supervision. 3) Joint supervision outperforms both SFT and RecA, demonstrating its effectiveness.

For image editing, we conduct analogous ablation studies and decompose the understanding objective into its individual components, where the setting without language or visual supervision corresponds to SFT on the same data. As reported in~\cref{tab:ablate_edit}, each supervision signal independently improves upon the SFT baseline, while their combination achieves the best performance on GEdit-Bench. These results mirror the image generation findings and further confirm the effectiveness and complementary benefits of language and visual understanding supervision.

\begin{wrapfigure}{r}{0.6\textwidth}
    \centering
    \vspace{-4mm}
    \includegraphics[width=\linewidth]{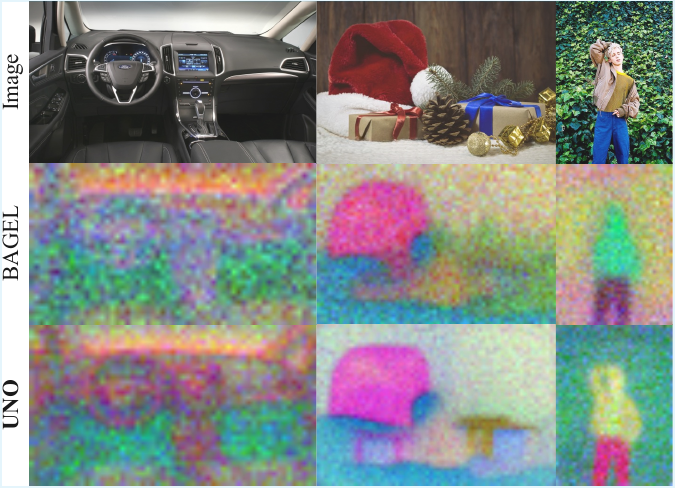}
    \caption{\textbf{Visualization of latent features}. We visualize the structure of latent features of the generation expert at highly noised timesteps. Empirically, we observe that understanding supervision improves latent structures, reduces noise while preserving better semantic information and details.}
    \vspace{-4mm}
    \label{fig:feature}
\end{wrapfigure}

\noindent\textbf{\ours \  improves generative features} Following prior feature-space analyses~\cite{kouzeliseq, leng2025repa}, we investigate the effect of \ours \  on generative representations by visualizing noised features from the generation expert. Specifically, we apply PCA to intermediate representations at early stages of denoising, as shown in~\cref{fig:feature}. When trained with generative supervision alone, representations exhibit substantial noise and weak semantic organization (\eg ambiguity in Col. 1), failing to preserve fine-grained visual details (\eg, missing gift box at the bottom-left in Col. 2, wrong pose in Col. 3). In contrast, \ours \  reduces noise and improves the discriminability of visual details, suggesting that understanding supervision encourages the model to form more semantically grounded and robust generative representations.

\begin{table*}[t]
    \centering
    \footnotesize 
    
    
    \begin{minipage}[t]{0.48\linewidth}
        \centering
        \caption{\textbf{Comparison between different supervision methods for post-training BAGEL}. For a fair comparison, we also implement RecA~\cite{xie2025reconstruction} on our training dataset, denoted by $^\dagger$.}
        \label{tab:post_train}
        \setlength{\tabcolsep}{1.5pt}
        \begin{tabular}{lccc}
        \toprule
        Model & GenEval2 & DPG-Bench & UniGenBench++ \\
        \midrule
        BAGEL & 71.7 & 84.03 & 61.53 \\
        w/ RecA & 74.9 & 85.29 & 63.74 \\
        \midrule
        w/ RecA$^\dagger$ & 74.3 & 85.37 & 64.12  \\
        w/ SFT & 73.6 & 84.91 & 63.02 \\
        \midrule
        w/ Language & 74.6 & 85.69 & 64.38 \\
        w/ Visual Und. & 74.0 & 85.31 & 63.60 \\
        w/ Joint & \textbf{75.1} & \textbf{86.12} & \textbf{65.03} \\
        \bottomrule
        \end{tabular}
    \end{minipage}\hfill
    \begin{minipage}[t]{0.48\linewidth}
        \centering
        \caption{\textbf{Effect of different supervision targets}. We decompose the joint supervision objective and ablate the individual contributions on image editing. Double cross mark denotes the SFT baseline trained without \ours. Lang. and Vis. denotes language and visual understanding supervision respectively.}
        \label{tab:ablate_edit}
        \setlength{\tabcolsep}{2.5pt}
        \begin{tabular}{cccccccc}
        \toprule
        \multirow{2}{*}{Lang.} & \multirow{2}{*}{Vis.} & \multicolumn{3}{l}{\ \ \ GEdit-Bench-EN} & \multicolumn{3}{l}{\ \ \ GEdit-Bench-CN} \\
        & & G\_SC      & G\_PQ     & \textbf{G\_O} & G\_SC      & G\_PQ     & \textbf{G\_O}     \\
        \midrule
        \xmark & \xmark & 7.61 & 7.44 & 7.00 & 7.63 & 7.42 & 6.96  \\
        \xmark & \cmark & 7.69 & 7.47 & 7.09 & 7.77 & 7.54 & 7.18 \\
        \cmark & \xmark & 7.66 & 7.47 & 7.09 & 7.69 & 7.54 & 7.11 \\
        \cmark & \cmark & \textbf{7.76} & \textbf{7.54} & \textbf{7.17} & \textbf{7.80} & \textbf{7.54} & \textbf{7.20} \\
        \bottomrule
        \end{tabular}
    \end{minipage}

    \vspace{1mm}

    \begin{minipage}[t]{0.48\linewidth}
        \centering
        \caption{\textbf{Effect of connector and supervision targets} for visual supervision. Sim.: cosine similarity. De.: Denoise. Default marked in \colorbox{defaultcolor}{gray}.}
        \label{tab:ablate_connector}
        \setlength{\tabcolsep}{1.5pt}
        \begin{tabular}{llccc}
        \toprule
        Arch. & Target & GenEval2 & DPG-Bench & UniGenBench++ \\
        \midrule
        \default{Identity} & \default{Sim.} & \default{\textbf{74.0}} & \default{\textbf{85.31}} & \default{63.60} \\
        ViT & Sim. & 73.7 & 85.21 & 63.47 \\
        MLP & De. & 73.7 & 85.09 & 63.41 \\
        DiT & De. & 73.8 & 85.09 & \textbf{63.82} \\
        \bottomrule
        \end{tabular}
    \end{minipage}\hfill
    \vspace{-2.5mm}
    \begin{minipage}[t]{0.48\linewidth}
        \centering
        \caption{\textbf{Effect of language augmentation}. Default in \colorbox{defaultcolor}{gray}.}
        \label{tab:ablate_semantic_aug}
        \setlength{\tabcolsep}{2.5pt}
        \begin{tabular}{ccccc}
        \toprule
        Aug. & GenEval2 & DPG-Bench & UniGenBench++ & Loss \\
        \midrule
        \xmark & 74.1 & 85.26 & 64.09 & $\sim$0.1 \\
        \default{\cmark} & \default{\textbf{74.6}} & \default{\textbf{85.69}} & \default{\textbf{64.38}} & \default{$\sim$0.7} \\
        \bottomrule
        \end{tabular}
    \end{minipage}
    \vspace{-4mm}

\end{table*}

\begin{wrapfigure}{r}{0.35\textwidth}
    \centering
    \vspace{-6mm}
    \includegraphics[width=\linewidth]{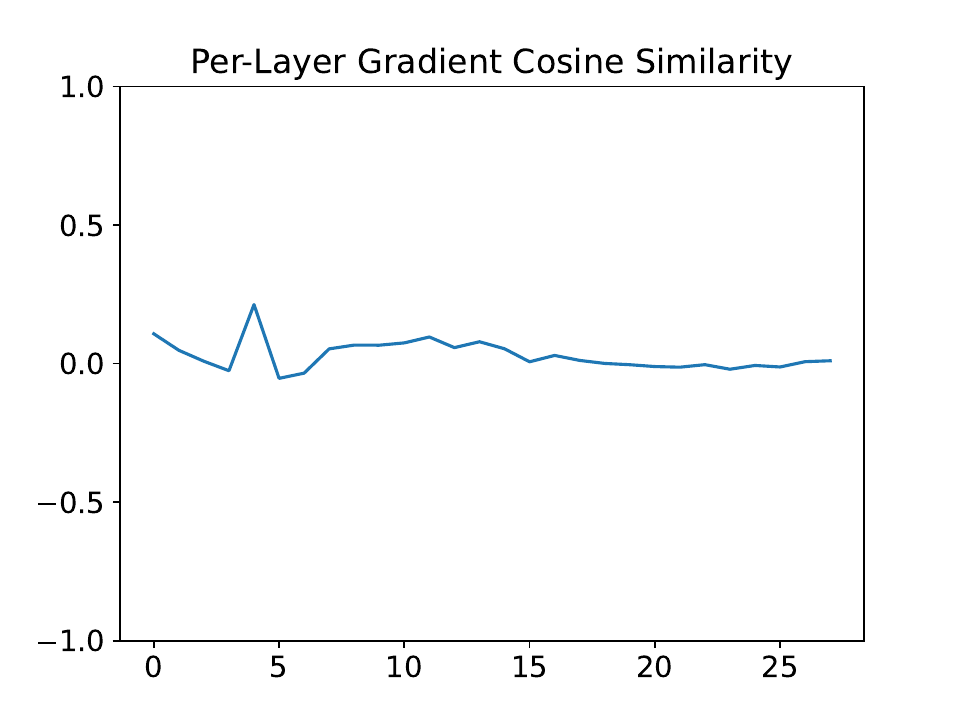}
    \caption{\textbf{Visualization of gradient similarity} between understanding and generation objectives.}
    \vspace{-4mm}
    \label{fig:grad_direction}
\end{wrapfigure}

\noindent\textbf{Optimization gradient directions} To investigate the effect of understanding supervision on optimization and gradient dynamics in generative training, we visualize the per-layer gradient directions induced by the understanding and generative objectives in \cref{fig:grad_direction}. We observe that gradients are largely orthogonal across most layers, with a subset exhibiting positive alignment. This suggests that, the understanding objective does not introduce optimization conflicts with denoising and instead provides complementary semantic guidance under \ours.

\noindent\textbf{Semantic augmentation for language supervision} We ablate the effect of semantic augmentation with different captions in ~\cref{tab:ablate_semantic_aug}. Empirically, we discover that augmentation effectively reduces leakage and enhances performance. We also discover that even though using the same prompt is prone to leakage, it can still improve generation.

\noindent\textbf{Vision connector design and more} We further study how different vision connector and supervision targets affect performance. We adopt a wide range of established approaches, including ViT-based similarity~\cite{pan2025transfer}, adaLN-MLP–style denoising~\cite{li2024autoregressive, team2025nextstep}, and DiT-based denoising reconstruction~\cite{wangreconstructive}. As summarized in~\cref{tab:ablate_connector}, we empirically observe that different architectures and targets results in similar performance for generation. Therefore, we adopt the most simple design using identity projection and cosine similarity. Further ablations are presented in \cref{sec:appendix_ablations} in the appendix.

\begin{wraptable}{r}{0.48\textwidth}
    \centering
    \footnotesize
    \setlength{\tabcolsep}{1.5pt}
    \caption{Effect of \ours \ on Show-o.}
    \label{tab:showo}
    \begin{tabular}{llccc}
    \toprule
    Model & GenEval2 & DPG-Bench & UniGenBench++ & MME \\
    \midrule
    Show-o & 58.57 & 80.04 & 50.39 & 1233 \\
    +\ours & \textbf{61.26} & \textbf{81.94} & \textbf{53.41} & \textbf{1238} \\
    \bottomrule
    \end{tabular}
    \vspace{-4mm}
\end{wraptable}

\noindent\textbf{General applicability} To further assess the generality of \ours, we conduct experiments on Show-o~\cite{xieshow}, an alternative unified model that integrates VLM with discrete diffusion. We train Show-o using the processed generation data with understanding data from LLaVA~\cite{liu2023visual} to preserve understanding capabilities. We report results across generation and understanding (MME~\cite{fumme}) benchmarks in ~\cref{tab:showo}. As demonstrated, \ours \ successfully improves upon the Show-o baseline across generation benchmarks while preserving understanding performance, demonstrating \ours's generality.

\begin{wraptable}{r}{0.48\textwidth}
    \centering
    \vspace{-4mm}
    \footnotesize
    \setlength{\tabcolsep}{1.5pt}
    \caption{Effect of \ours \ and CoT on BAGEL.}
    \label{tab:effect_cot}
    \begin{tabular}{cc|ccc}
    \toprule
    CoT & \ours & GenEval2 & DPG-Bench & UniGenBench++ \\
    \midrule
    \xmark & \xmark & 69.6 & 84.03 & 61.53 \\
    \xmark & \cmark & 72.0 & \textbf{86.12} & \textbf{65.03} \\
    \cmark & \xmark & 71.7 & 83.69 & 61.36 \\
    \cmark & \cmark & \textbf{75.1} & 85.33 & 64.95 \\
    \bottomrule
    \end{tabular}
    \vspace{-2mm}
\end{wraptable}

\noindent\textbf{\ours\ and CoT} Self-CoT is a commonly adopted strategy for improving generation performance in unified models. We conduct a systematic analysis of the interplay between CoT and \ours. Our results in \cref{tab:effect_cot} show that \ours\ consistently outperforms CoT alone, while also serving as a complementary mechanism that further enhances CoT-based generation when combined.

\begin{wraptable}{r}{0.48\textwidth}
    \centering
    \vspace{-4mm}
    \footnotesize
    \setlength{\tabcolsep}{1.5pt}
    \caption{\textbf{Evaluation on WISE}. $^*$ indicates additional training with augmented knowledge data.}
    \label{tab:wise}
    \begin{tabular}{lccccccc}
    \toprule
    Model & Cult. & Time & Spa. & Bio. & Phy. & Chem. & Ovr. \\
    \midrule
    BAGEL & 0.44 & 0.55 & 0.68 & 0.44 & 0.60 & 0.39 & 0.52 \\
    +\ours & 0.44 & 0.52 & 0.63 & 0.45 & 0.57 & 0.42 & 0.51 \\
    +\ours$^*$ & 0.49 & 0.56 & 0.64 & 0.52 & 0.67 & 0.46 & \textbf{0.56} \\
    \bottomrule
    \end{tabular}
    \vspace{-2mm}
\end{wraptable}

\label{sec:limitations}

\noindent\textbf{Limitations and future work} \ours\ is a general-purpose framework that does not leverage specialized data for vertical domains. Therefore, on tasks requiring such capabilities (\eg, knowledge retrieval in WISE~\cite{niu2025wise}), it remains largely orthogonal. Preliminary results indicate these capabilities are complementary and do not conflict with our design. To verify this, we construct 50k distilled knowledge-oriented samples and continue training \ours\ for 1k steps, yielding consistent gains (\cref{tab:wise}). We will further explore improving on these specialized domains in future work.
\section{Conclusion}
\label{sec:conclusion}

In this work, we present \ours, a preliminary exploration into supervising generation directly with understanding in unified multimodal models. Through designing a light-weight post-training framework that strengthens gradient flow from understanding to generation via a combination of complementary objectives, we demonstrate that understanding can serve as effective catalyst to enhance generation. Extensive experiments across different generative tasks validate the effectiveness of \ours. Our findings show promising possibilities for more integrated unified models with deeper synergy.

\section*{Acknowledgements}

This work is supported in part by the National Key R\&D Program of China under Grant 2024YFB4708200, the National Natural Science Foundation of China under Grants U24B20173, U2541227 and 62625604, the Lenovo Academic Collaboration Project, and the Scientific and Techno-logical innovation action plan of Shanghai Science and Technology Committee (No.22511101502). We would also like to extend our deepest appreciation to Jiayi Guo, Yifan Pu and Xu Zhang for insightful discussions.

\bibliographystyle{plain}
\bibliography{main}


\newpage
\appendix
\section{Training Settings}

\label{sec:training_param}

We present the detailed training parameters for training image generation and editing tasks in ~\cref{tab:training_param}. All experiments are trained with compute equivalent to approximately 32 A100 GPU days.

\begin{table}[h]
\centering
\footnotesize
\caption{Detailed hyper-parameters for post-training BAGEL on image generation and editing tasks.}
\begin{tabular}{l|c|c}
Hyper-parameter & Image Generation & Image Editing \\
\midrule
Iterations &  5K & 5K \\
\#Samples & $\sim$2M & $\sim$750K \\
Learning Rate & 0.0001 & 0.0001 \\
Schedule & Cosine & Constant \\
Weight Decay & 0.0 & 0.0 \\
$\lambda_1$ & 0.1 & 0.1 \\
$\lambda_2$ & 0.2 & 0.2 \\
EMA ratio & 0.9999 & 0.9999 \\
\#Metaqueries & 256 & 256 \\
Grad Norm Clip & 1.0 & 1.0 \\
Timestep Shift & 4.0 & 4.0 \\
Resolution & (512, 1024) & (512, 1024)
\end{tabular}
\label{tab:training_param}
\end{table}

\section{Training Sample Statistics}

\label{sec:training_statistics}

We report the average token count for each component in the packed training samples of \ours. As shown in \cref{tab:data_statistics}, \ours\ introduces only a marginal token overhead (text captions and meta-query tokens) compared to standard text-to-image training (text conditions and image tokens).

\begin{table}[h]
\centering
\footnotesize
\caption{Average token count for each component when training \ours.}
\begin{tabular}{c|c|c|c}
Text Cond. & Img & Text Cap. & Metaquery \\
\midrule
171 & 3362 & 80 & 256
\end{tabular}
\label{tab:data_statistics}
\end{table}

\section{Image Generation Training Data}

\label{sec:training_data}

We curate our dataset from a diverse set of open-source resources, including LAION~\cite{schuhmann2022laion}, JourneyDB~\cite{sun2023journeydb}, and OpenImages~\cite{kuznetsova2020open}, etc. The collected data spans a wide range of categories, encompassing natural, synthetic, and design imagery, with rich subcategories such as objects, landscapes, human subjects, animals, plants, food, indoor scenes, artistic designs, and sports. Following data collection, we apply a systematic filtering pipeline based on resolution (excluding images below 512 pixels), aesthetic quality, visual clarity, color saturation, and safety considerations. Finally, we employ an internal captioning model to produce detailed, dense textual descriptions, yielding prompts with an average length of approximately 200 tokens.

For the recaptioning pipeline, we adopt a simple and straightforward approach, prompting Qwen2.5-VL-7B with the target image, and the system prompt "You are an ai captioning assistant, please describe this image in detail. Be absolutely accurate with your caption, do not imagine, hallucinate or hint at contents that is NOT present in the image."

\begin{figure*}[t]
    \centering
    \includegraphics[width=\linewidth]{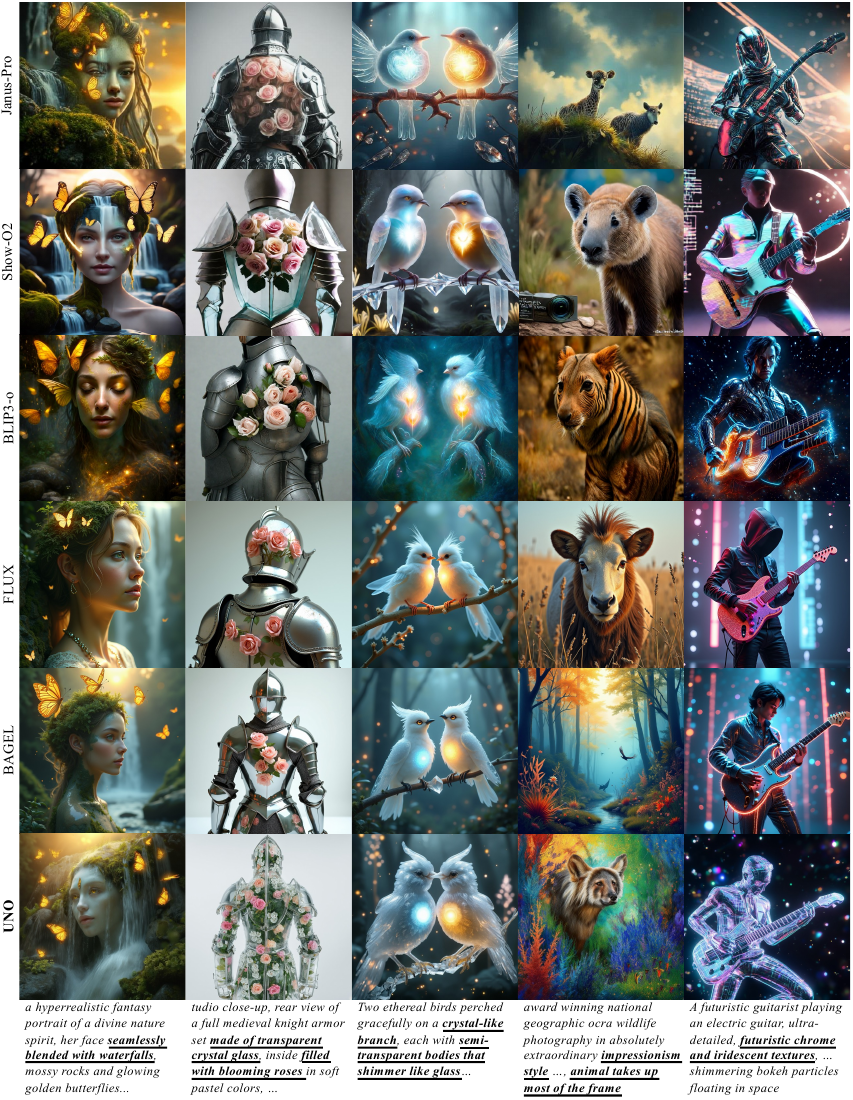}
    \caption{\textbf{More complete qualitative comparisons on image generation} between \ours \ and competitive generation and unified model baselines. }
    \label{fig:cmp_gen_full}
    \vspace{-4mm}
\end{figure*}

\section{Further Ablations}

\label{sec:appendix_ablations}

\noindent{\textbf{Number of Metaqueries}} We study the impact of varying the number of metaqueries used for visual supervision, with results reported in \cref{tab:ablate_num_metaquery}. As shown, increasing the number of metaqueries beyond 256 leads to a degradation in performance. This trend suggests that visual understanding supervision at a resolution of $224 \times 224$ is sufficient to capture the necessary semantic details for effectively supervising the generation process.

\noindent{\textbf{Unfreezing the Understanding Expert}} We ablate the effect of jointly training the understanding expert, with results summarized in \cref{tab:ablate_train_und}. We find that unfreezing and fine-tuning the understanding expert does not yield consistent improvements in generation performance. Moreover, optimizing the understanding expert solely with the proxy objective may degrade its performance on standard understanding benchmarks, which can in turn weaken the quality of the supervision signals provided to the generation expert.

\noindent\textbf{Effect of masking condition prompts} We investigate the effect of masking the condition prompts and present results in \cref{tab:effect_masking}. We observe that without masking, information leakage mostly undermines the effect of understanding supervision.

\noindent\textbf{Effect of causal prediction for visual supervision} We investigate the effect of masking the condition prompts and present results in \cref{tab:effect_causal}. We observe that predicting the metaquery tokens in causal or bidirectional order does not significantly affect performance.

\begin{table}[htbp]
    \centering
    \footnotesize
    
    \begin{minipage}{0.48\linewidth}
        \centering
        \setlength{\tabcolsep}{2.5pt}
        \captionof{table}{
        \textbf{Effect of different number of metaqueries}. Default marked in \colorbox{defaultcolor}{gray}.}
        \begin{tabular}{cccc}
        \toprule
        \#Metaqueries & GenEval2 & DPG-Bench & UniGenBench++ \\
        \midrule
        \default{256} & \default{74.0} & \default{85.31} & \default{63.60} \\
        576 & 73.4 & 85.08 & 63.36 \\
        \bottomrule
        \end{tabular}
        \label{tab:ablate_num_metaquery}
    \end{minipage}\hfill
    \begin{minipage}{0.48\linewidth}
        \centering
        \setlength{\tabcolsep}{1.5pt}
        \captionof{table}{
        \textbf{Effect of unfreezing the understanding expert}. Default marked in \colorbox{defaultcolor}{gray}.}
        \begin{tabular}{cccc}
        \toprule
        Train Und. & GenEval2 & DPG-Bench & UniGenBench++ \\
        \midrule
        \default{\xmark} & \default{75.1} & \default{86.12} & \default{65.03} \\
        \cmark & 75.8 & 84.62 & 64.36 \\
        \bottomrule
        \end{tabular}
        \label{tab:ablate_train_und}
    \end{minipage}
    
    \vspace{1em} 
    
    
    \begin{minipage}{0.48\linewidth}
        \centering
        \setlength{\tabcolsep}{2.5pt}
        \captionof{table}{
        \textbf{Effect of masking conditional prompts}. Default marked in \colorbox{defaultcolor}{gray}.}
        \begin{tabular}{cccc}
        \toprule
        Mask & GenEval2 & DPG-Bench & UniGenBench++ \\
        \midrule
        \xmark & 73.4 & 84.97 & 62.76 \\
        \default{\cmark} & \default{75.1} & \default{86.12} & \default{65.03} \\
        \bottomrule
        \end{tabular}
        \label{tab:effect_masking}
    \end{minipage}\hfill
    \begin{minipage}{0.48\linewidth}
        \centering
        \setlength{\tabcolsep}{1.5pt}
        \captionof{table}{
        \textbf{Effect of prediction order for visual supervision}. Default marked in \colorbox{defaultcolor}{gray}.}
        \begin{tabular}{cccc}
        \toprule
        & GenEval2 & DPG-Bench & UniGenBench++ \\
        \midrule
        \default{Causal} & \default{75.1} & \default{86.12} & \default{65.03} \\
        Bidirectional & 75.3 & 85.98 & 65.14 \\
        \bottomrule
        \end{tabular}
        \label{tab:effect_causal}
    \end{minipage}
    
\end{table}

\section{More Visualization Comparisons}

\label{sec:more_visualization}

We provide a more comprehensive qualitative comparison in \cref{fig:cmp_gen_full}, including representative generative and unified baselines such as FLUX~\cite{flux}, Janus-Pro~\cite{chen2025janus}, BLIP3-o~\cite{chen2025blip3}, Show-o2~\cite{xie2025show}, and BAGEL~\cite{deng2025emerging}.

\section{Editing Evaluations}

We present a more comprehensive evaluation on image editing on ImgEdit~\cite{yeimgedit} and KRIS-Bench~\cite{wukris}, in \cref{tab:imgedit} and \cref{tab:kris} respectively. 

\begin{table}[htbp]
    \centering
    \footnotesize
    \setlength{\tabcolsep}{2.5pt}
    
    \caption{Quantitative comparisons on ImgEdit.}
    \label{tab:imgedit}
    \begin{tabular}{lcccccccccc}
    \toprule
    Model & Bg. & Sty. & Adj. & Ext. & Rem. & Rep. & Add. & Cmp. & Act. & Ovr. \\
    \midrule
    FLUX-Kontext & 3.89 & 4.62 & 3.69 & 1.81 & 2.97 & 4.20 & 3.80 & 3.00 & 4.18 & 3.6 \\
    BAGEL & 3.38 & 4.53 & 3.58 & 1.49 & 3.15 & 3.82 & 3.71 & 2.64 & 4.21 & 3.38 \\
    \textbf{+\ours} & 4.01 & 4.83 & 4.14 & 2.2 & 3.58 & 4.40 & 4.10 & 3.06 & 4.32 & \textbf{3.85} \\
    \bottomrule
    \end{tabular}
    
    \vspace{3mm} 
    
    \caption{Quantitative comparisons on KRIS-Bench.}
    \label{tab:kris}
    \begin{tabular}{lcccccccc}
    \toprule
    Model & AP & SP & TP & SS & NS & LP & ID & Ovr. \\
    \midrule
    BAGEL-think & 67.42 & 68.33 & 58.67 & 63.55 & 61.40 & 48.12 & 50.22 & 60.18 \\ 
    \textbf{+\ours} & 73.09 & 72.25 & 59.18 & 64.65 & 61.51 & 48.00 & 55.67 & \textbf{62.57} \\
    \bottomrule
    \end{tabular}
    \vspace{-4mm}
\end{table}

\begin{table}[ht!]
\centering
\footnotesize
\begin{tabular}{l|>{\RaggedRight\arraybackslash}p{6cm}}
Image & Prompt \\
\midrule
\scriptsize{Row 1, Col 1} & \scriptsize{In a surrealist oil painting, the body and horns of a giraffe are composed of bright stars, and many butterflies, also composed of nebulae, are resting on its horns.} \\ \hline
\scriptsize{Row 1, Col 2} & \scriptsize{An open book the size of a mountain range, pages crumbling into sandy dunes with ancient scripts glowing, a wanderer climbing the spine amid a sandstorm, warm amber lights and swirling dust particles} \\ \hline 
\scriptsize{Row 1, Col 3} & \scriptsize{An astronaut is riding on the back of a giant dragon composed of brilliant stardust and diffuse nebulae; the dragon's body is translucent and iridescent. They are shuttling at high speed through Saturn's magnificent rings, with countless glowing ice crystals and rock particles dancing around. In the distant background is Saturn's huge orange-yellow sphere. The entire scene presents a magnificent oil painting texture, with heavy brushstrokes, rich and saturated colors, and strong contrast between light and shadow. The light from the stardust illuminates the astronaut's helmet and deep space, filled with an epic and grand momentum.} \\ \hline
\scriptsize{Row 1, Col 4} & \scriptsize{A fantasy city made of huge candies and biscuits, all the buildings and streets in the city present a sweet and seductive texture.} \\ \hline
\scriptsize{Row 1, Col 5} & \scriptsize{A crystal clear crystal ball is wrapped in a collapsing Great Wall, surrounded by fine snowflakes.} \\ \hline
\scriptsize{Row 2, Col 1} & \scriptsize{Draw a grand piano made of transparent crystal. When the keys are pressed, swarms of glowing butterflies fly out instead of musical notes.} \\ \hline
\scriptsize{Row 2, Col 2} & \scriptsize{a miniature house inside an apple and you can see little, cute pandas living inside the apple, eating dinner at the dinner table, rendered hyper-realistic with strong depth of field to make it seem very miniature} \\ \hline
\scriptsize{Row 2, Col 3} & \scriptsize{A glowing giant whale whose body is made of a nebula has a coral reef made of broken planets growing on its back.} \\ \hline
\scriptsize{Row 2, Col 4} & \scriptsize{A realistic blurred photo of a vibrant city street at golden hour. In the foreground, a glowing neon line art of a joyful boy riding a skateboard. He is outlined in thick glowing white lines, wearing a hoodie with sketchy lightning bolts and stars around him. his shoes glow as he balances with one foot up. Graffiti-style hearts and sparkles trail behind his like motion lines.} \\ \hline
\scriptsize{Row 2, Col 5} & \scriptsize{A mesmerizing hyper-realistic digital sketch by artist PJ, featuring an egg cracked open to reveal the miniature, intricate city of London within.The eggshell fragments, both upright and flat, create a chaotic yet harmonious composition against the smooth, white background. The vibrant cityscape, with Tower Bridge, River Thames, Big Ben, Buckingham Palace, and the most charming Londonian landmarks, contrasts beautifully with the muted tones of the kitchen counter in the background. The artist's use of textures and shadows adds depth, while the vibrant colors and gritty style make this scene truly captivating, kitchen background} \\
\end{tabular}
\caption{Detailed prompt list for qualitative image generation results in \cref{fig:visualize_gen}.}
\label{tab:full_prompt}
\end{table}

\section{Evaluation Robustness}

\label{sec:mean_std}

\begin{wraptable}{r}{0.5\textwidth}
    \centering
    \vspace{-4mm}
    \caption{Mean and variance of \ours.}
    \label{tab:mean_std}
    \footnotesize
    \setlength{\tabcolsep}{2.5pt}
    \begin{tabular}{lccc}
    \toprule
    Statistics & GenEval2 & DPG-Bench & UniGenBench++ \\
    \midrule
    Mean & 75.1 & 86.12 & 65.14 \\
    Variance & 0.034 & 0.024 & 0.017 \\
    \bottomrule
    \end{tabular}
\end{wraptable}

We report the mean and variance of the main evaluation with 4 random seeds in \cref{tab:mean_std}, demonstrating the statistical robustness of our evaluation.

\section{Detailed Prompt List}

We illustrate the detailed prompts for generated images shown in \cref{fig:visualize_gen} in \cref{tab:full_prompt}.

\section{Societal Impact and Safeguards}

\label{sec:societal_impact}

This paper aims at foundational research and not tied to particular applications, let alone deployments. However, we point out that a general improvement in the quality of generative models proposed in this paper could be used to generate Deepfakes for disinformation. The released models should be used in accordance with the Apache-2.0 license and accompanying terms of use, which are intended to mitigate potential abuse.


\end{document}